\documentclass[conference]{IEEEtran}

\IEEEoverridecommandlockouts

\usepackage{graphicx}
\usepackage{amsmath}
\usepackage{amssymb}
\usepackage{algorithm}
\usepackage{algorithmic}
\usepackage{booktabs}
\usepackage{multirow}
\usepackage{subcaption}
\usepackage{hyperref}
\usepackage{cite}
\usepackage[T1]{fontenc}

\begin{document}

\title{Local-Global Feature Fusion for Subject-Independent EEG Emotion Recognition}

\author{\IEEEauthorblockN{ Zheng Zhou}
\IEEEauthorblockA{\textit{Computer Science and Technology} \\
\textit{Beijing Jiaotong University}\\
Beijing, China \\
zhengzhou@bjtu.edu.cn}
\and
\IEEEauthorblockN{ Isabella McEvoy}
\IEEEauthorblockA{\textit{Applied Computer Science} \\
\textit{The University of Winnipeg}\\
Winnipeg, Canada \\
mcevoy-i@webmail.uwinnipeg.ca}
\and
\IEEEauthorblockN{ Camilo E. Valderrama}
\IEEEauthorblockA{\textit{Applied Computer Science} \\
\textit{The University of Winnipeg}\\
Winnipeg, Canada \\
c.valderrama@uwinnipeg.ca}
}


\maketitle

\thispagestyle{empty}
\pagestyle{empty}

\begin{abstract}
Subject-independent EEG emotion recognition is challenged by pronounced inter-subject variability and the difficulty of learning robust representations from short, noisy recordings. To address this, we propose a fusion framework that integrates (i) local, channel-wise descriptors and (ii) global, trial-level descriptors, improving cross-subject generalization on the SEED-VII dataset. Local representations are formed per channel by concatenating differential entropy with graph-theoretic features, while global representations summarize time-domain, spectral, and complexity characteristics at the trial level. These representations are fused in a dual-branch transformer with attention-based fusion and domain-adversarial regularization, with samples filtered by an intensity threshold. Experiments under a leave-one-subject-out protocol demonstrate that the proposed method consistently outperforms single-view and classical baselines, achieving approximately 40\% mean accuracy in 7-class subject-independent emotion recognition. The code has been released at \url{https://github.com/Danielz-z/LGF-EEG-Emotion}. 
\end{abstract}

\section{INTRODUCTION}

Emotion recognition from electroencephalography (EEG) has become an essential component in affective computing, mental-state monitoring, and brain-computer interaction (BCI)~\cite{alarcao2019survey,jenke2014survey}.
Compared with facial expressions or peripheral physiological signals, EEG provides a more direct measure of neural activity and is therefore less susceptible to external manipulation, cultural bias, or voluntary control.

Early studies have shown that spectral features extracted from EEG channels can effectively characterize emotion-related neural patterns~\cite{zheng2015tamd,zheng2014icme}.
Despite these advantages, building reliable EEG-based emotion recognition systems remains challenging due to substantial inter-subject variability in brain dynamics, recording conditions, and affective responses.
Even when temporal smoothing or stability modeling techniques are applied~\cite{shi2010lds,zheng2019tac}, models trained on one group of individuals often struggle to generalize to unseen subjects~\cite{apicella2024generalization}, highlighting the need for robust representations that capture invariant neural patterns while preserving emotionally discriminative information. To mitigate subject-related distribution shifts, domain adaptation and adversarial learning strategies have been explored in EEG-based emotion recognition~\cite{ganin2016dann,ma2023tnsre,shen2023contrastive}, yet robust subject-independent performance remains challenging.

To address the high inter-subject variability inherent in EEG signals, recent studies have shown that explicitly modeling global spatial dependencies among EEG channels can improve emotion recognition performance by capturing inter-channel relationships \cite{song2020dgcnn}. However, most existing approaches predominantly emphasize either \emph{local}, channel-wise representations or \emph{global} representations in isolation, thereby missing the opportunity to leverage their complementary strengths. Local features often fail to capture cross-channel interactions that reflect integrative, network-level neural processing, whereas global representations alone may obscure fine-grained temporal and scale-dependent information present at individual electrodes. Therefore, the joint modeling of local and global EEG features for robust emotion recognition remains an underexplored research direction.

Building on this motivation, we propose a \emph{local–global connectivity fusion framework} for subject-independent EEG emotion recognition on the SEED-VII dataset~\cite{seedvii}. Specifically, From 62 EEG channels, we construct \emph{local} channel-wise representations by concatenating differential entropy (DE) features with graph-theoretic connectivity descriptors, resulting in a $62 \times (5{+}4)$ feature map that is flattened into a 558-dimensional EEG vector. In parallel, \emph{global} descriptors are summarized as a 25-dimensional, trial-level auxiliary vector capturing multifractal, temporal, and spectral characteristics aggregated across channels. These two modalities are jointly processed by a dual-branch transformer backbone (MAET) under a leave-one-subject-out (LOSO) evaluation protocol, incorporating subject-wise normalization and domain-adversarial regularization to mitigate cross-subject distribution shifts.

The contributions of this work are threefold:
\begin{enumerate}
    \item We propose a unified local-global EEG representation that jointly models channel-level neural dynamics and node-level functional connectivity for subject-independent emotion recognition.
    \item We develop a dual-branch transformer-based fusion framework that integrates heterogeneous EEG and auxiliary features while alleviating cross-subject distribution shifts.
    \item Extensive experiments on the SEED-VII dataset under a strict LOSO protocol demonstrate that the proposed method consistently outperforms local-only, global-only, and classical machine learning baselines in 7-class emotion recognition.
\end{enumerate}

\section{RELATED WORK}

Most existing EEG-based emotion recognition approaches primarily rely on \emph{local} channel-wise signal features. Among these local descriptors, spectral features have proven effective in modeling oscillatory patterns associated with emotional states. In particular, studies on the SEED-family datasets have demonstrated that DE features, especially when smoothed using linear dynamical systems (LDS), provide a stable and physiologically meaningful representation of EEG responses.

Beyond spectral analysis, functional connectivity and graph-theoretic measures have been widely adopted to characterize large-scale interaction patterns across brain regions~\cite{bullmore2009natrev,rubinov2010neuroimage}. These approaches capture network-level coordination and have shown relevance for affective and cognitive processing by modeling inter-regional relationships rather than isolated channel activity.

More recently, the introduction of self-attention mechanisms has enabled transformer architectures to model long-range dependencies and contextual interactions in sequential data~\cite{vaswani2017attention}. Several studies have successfully applied transformer-based or attention-enhanced models to EEG emotion recognition, demonstrating their ability to capture temporal dependencies and complex feature interactions~\cite{xiao2021attention4d,hu2022cnn_bilstm_mhsa}.

Despite these advances, approaches relying solely on local features often fail to capture cross-channel dependencies that reflect integrative, network-level neural processing. Conversely, models using global features may ignore fine-grained temporal and frequency-specific information present at individual electrodes. Moreover, the effectiveness of transformer-based models strongly depends on the structure and complementarity of their input representations; inadequately organized features can hinder cross-subject generalization. These limitations motivate fusion strategies that jointly exploit local neural dynamics and global coordination patterns for robust, subject-independent EEG emotion recognition. Accordingly, our work advances this direction by integrating local channel-wise descriptors with global trial-level representations within a unified learning framework.



\section{METHODS}
\subsection{Dataset: SEED-VII}

The SEED-VII dataset~\cite{seedvii} contains EEG recordings associated with seven discrete emotional states: \textit{happy}, \textit{sad}, \textit{fear}, \textit{disgust}, \textit{surprise}, \textit{anger}, and \textit{neutral}. The data were collected from 20 participants (10 male and 10 female), each of whom completed four recording sessions. During each session, participants viewed 20 emotion-eliciting movie clips, resulting in a total of 80 clips per participant across all sessions.

\subsection{Preprocessing}

The raw EEG signals were first band-pass filtered between 0.1 and 70 Hz to suppress slow drifts and high-frequency noise. A 50 Hz notch filter was then applied to remove power-line interference. To reduce computational complexity, the signals were downsampled to 200 Hz, which, according to the Nyquist theorem, preserves spectral information below 100 Hz. Finally, to enhance temporal dynamics and mitigate noise, the EEG signals were smoothed using a Linear Dynamical System (LDS), following the procedure described in~\cite{shi2010lds}.

\begin{figure*}[t]
    \centering
    \includegraphics[width=\textwidth]{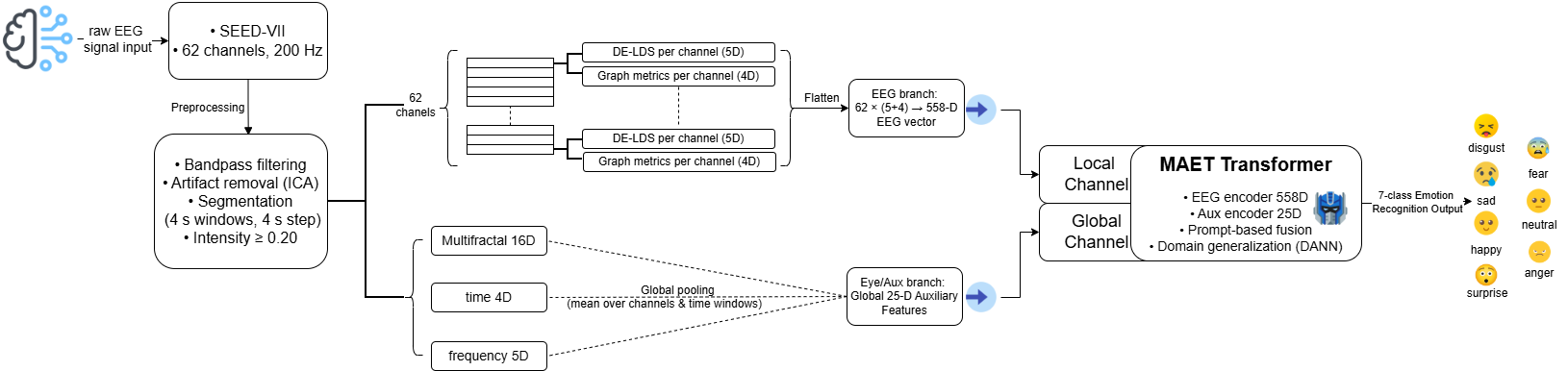}
    \caption{Overall workflow of the proposed pipeline on the SEED-VII dataset. Preprocessed EEG signals were used to extract local channel-wise and global trial-level features, which were subsequently integrated via an MAET-based fusion model and evaluated under a LOSO protocol.}
    \label{fig:workflow}
\end{figure*}

\subsection{Overall Framework}

Fig.~\ref{fig:workflow} illustrates the processing pipeline. From each trial, two types of feature representations were extracted: 
(1) a channel-wise EEG feature vector encoding local neural spectral and connectivity information, and 
(2) a trial-level descriptor summarizing global characteristics. These two modalities were fed into a dual-branch transformer backbone with attention-based fusion and domain-adversarial regularization.

\begin{figure*}[t]
    \centering
    \includegraphics[width=\textwidth]{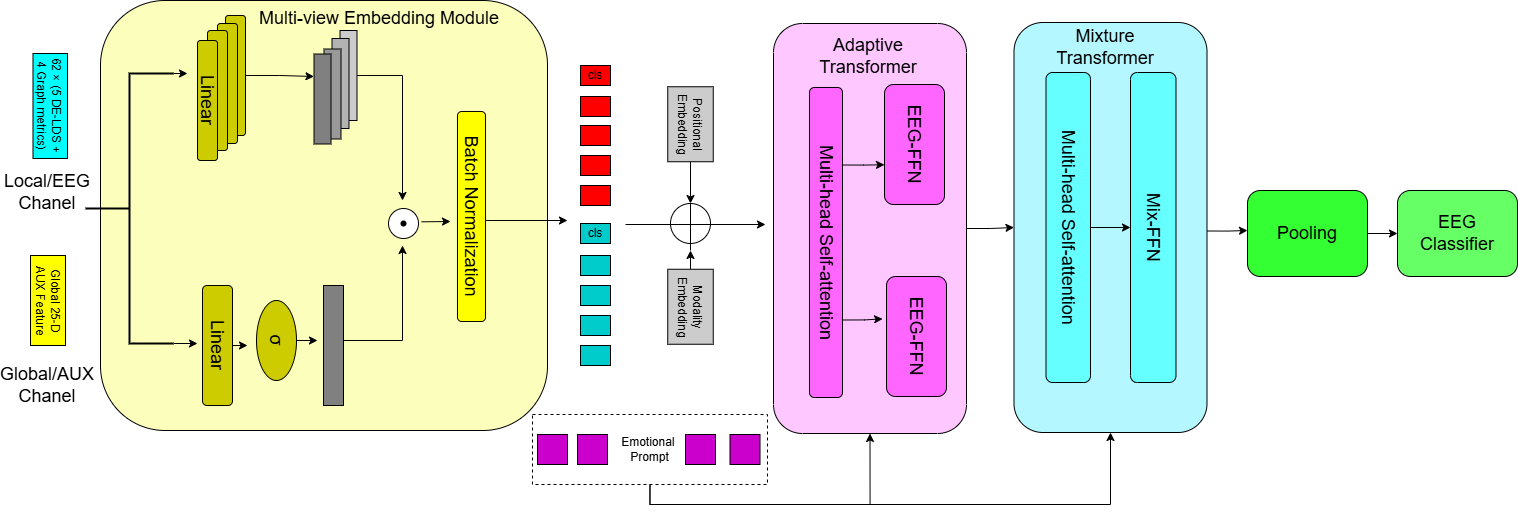}
    \caption{Architecture of the MAET backbone: multi-view embeddings for the EEG feature vector (local dynamics and node-level connectivity) and the auxiliary feature vector, followed by transformer blocks and a single emotion classification head.}
    \label{fig:maet}
\end{figure*}

\subsubsection{Local Feature}
Local EEG features were computed in a channel-wise manner.
Specifically, DE features and node-level
graph-theoretic connectivity descriptors were extracted for each EEG channel.

\paragraph{Differential Entropy}
DE features were extracted from each EEG channel. For each trial, the short-time power spectral density (PSD) was estimated using a Hann-windowed short-time Fourier transform (STFT). Given the EEG signal $x_c(t)$ from channel $c$, the PSD was computed as
\[
    P_c(f) = \frac{1}{T}\left|\mathcal{F}\{x_c(t)\}\right|^2,
\]
and following \cite{liu2021comparing}, the DE for channel $c$ in frequency band $b$ was approximated as:
\[
    \mathrm{DE}_c^{(b)} = \log_2 \left(100 \times P_c(f) \right).
\]
The resulting DE feature tensor for each trial was denoted as $\mathbf{X} \in \mathbb{R}^{ 62 \times 5}$, where 62 corresponded to the number of EEG channels and 5 represented the five canonical EEG frequency bands: delta, theta, alpha, beta, and gamma ($\delta$, $\theta$, $\alpha$, $\beta$, and $\gamma$).

\paragraph{Graph-theory features}
To characterize large-scale coordination patterns across channels, we constructed a correlation-based functional connectivity matrix for each trial. The correlation matrix was computed as
\[
    C = \mathrm{ReLU}\!\left(\mathrm{corrcoef}(X)\right),
\]
where negative correlations were removed and self-connections were discarded.

From $C$, four node-level graph-theoretic measures were computed for each channel, including weighted degree, local clustering coefficient, a betweenness centrality, and PageRank. These graph-theoretic measures are widely used to characterize functional brain networks and provide meaningful descriptors of large-scale neural coordination~\cite{bullmore2009natrev,rubinov2010neuroimage}.

\paragraph{Weighted Degree}
The weighted degree (strength) of node $i$ was defined as the sum of the weights of all edges connected to the node:
\begin{equation}
k_i^{(w)} = \sum_{j \in V} c_{ij}.
\end{equation}

\paragraph{Local Clustering Coefficient}
The local clustering coefficient of node $i$ quantified the density of connections among its neighbors and was defined as:
\begin{equation}
C_i = \frac{2 e_i}{k_i (k_i - 1)},
\end{equation}
where $k_i$ denoted the (unweighted) degree of node $i$ and $e_i$ denoted the number of edges between its neighbors.

\paragraph{Betweenness Centrality}
Betweenness centrality of node $i$ was defined as the fraction of shortest paths between all node pairs that passed through node $i$:
\begin{equation}
B_i = \sum_{\substack{s \neq i \neq t \\ s,t \in V}} \frac{\sigma_{st}(i)}{\sigma_{st}},
\end{equation}
where $\sigma_{st}$ denoted the total number of shortest paths between nodes $s$ and $t$, and $\sigma_{st}(i)$ denoted the number of those paths that passed through node $i$.

\paragraph{PageRank}
PageRank assigned each node $i$ a score based on the scores of nodes linking to it and was defined recursively as:
\begin{equation}
PR_i = \frac{1-d}{|V|} + d \sum_{j \in \mathcal{N}_i^{\text{in}}} \frac{PR_j}{k_j^{\text{out}}},
\end{equation}
where $d \in (0,1)$ denoted the damping factor, $\mathcal{N}_i^{\text{in}}$ denoted the set of nodes with edges directed toward node $i$, and $k_j^{\text{out}}$ denoted the out-degree of node $j$. PageRank values were normalized such that $\sum_i PR_i = 1$.

The resulting graph-based feature matrix was
\[
    \mathbf{F}_{\text{conn}} \in \mathbb{R}^{ 62 \times 4}.
\]

\subsubsection{Global Feature}

A global feature vector was introduced as a second modality. Global descriptors were obtained by computing time-domain, spectral, and multifractal features for each EEG channel, followed by average pooling across channels to form a compact trial-level representation.

Formally, for any per-channel feature matrix $\mathbf{F}\in\mathbb{R}^{62\times d}$, we applied channel-wise average pooling to obtain the trial-level descriptor $\bar{\mathbf{f}}=\frac{1}{62}\sum_{c=1}^{62}\mathbf{F}_{c,:}\in\mathbb{R}^{d}$.

The resulting global feature vector was 25-dimensional and comprised descriptors characterizing statistical time-domain measures, spectral band-power features capturing global frequency characteristics, and scale-dependent signal irregularity. This design enables the encoding of complementary global information beyond channel-wise EEG features while maintaining a compact representation.

\paragraph{Time-domain features}
Time-domain features were extracted to characterize the overall amplitude variability of EEG signals within each trial. For each channel $c$, given the EEG signal $x_c(t)$ of length $T$, we computed the first four statistical moments: mean, standard deviation, skewness, and kurtosis. The resulting per-channel time-domain feature matrix was
\[
\mathbf{F}_{\mathrm{time}}\in\mathbb{R}^{62\times4}.
\]


\paragraph{Spectral features}
Spectral features were extracted to summarize the global frequency characteristics of EEG activity. For each channel $c$, the PSD $P_c(f)$ was estimated, and band-wise spectral power was computed by integrating the PSD over the canonical EEG frequency bands:
\[
    P_c^{(b)} = \int_{f \in \mathcal{B}_b} P_c(f)\, df ,
\]
where $\mathcal{B}_b \in \{\delta, \theta, \alpha, \beta, \gamma\}$ denotes the $b$-th frequency band. The resulting per-channel spectral feature matrix was
\[
\mathbf{F}_{\mathrm{spec}}\in\mathbb{R}^{62\times5}.
\]


\paragraph{Multifractal Analysis Based on Wavelet Leaders}

Multifractal analysis provides a principled framework for modeling nonlinear and scale-dependent dynamics in complex signals~\cite{chhabra1989prl}, and has been widely adopted for characterizing EEG and other physiological time series~\cite{ihlen2012frontiers,zorick2013plosone,franca2018fractal}.
Recent studies further suggest that EEG multifractal properties are closely related to cognitive and affective processes~\cite{zorick2020jcn}. 

To characterize scale-dependent irregularity and nonlinear dynamics in EEG signals, we extracted multifractal descriptors using a wavelet-leader-based multifractal formalism combined with the Chhabra--Jensen direct method~\cite{chhabra1989prl,ihlen2012frontiers,franca2018fractal},
which enables robust estimation of multifractal properties from finite-length, noisy physiological time series.

Let $x_c(t)\in\mathbb{R}^{T}$ denote the EEG time series from channel $c$. A discrete wavelet transform was first applied to obtain wavelet coefficients across dyadic scales $j$, and wavelet leaders $L(j,k)$ were constructed as local suprema of wavelet coefficients within spatial neighborhoods, capturing the most singular structures at each scale.

For a set of moment orders $q\in[q_{\min},q_{\max}]$, the structure functions were computed as
\[
S(q,j)=
\begin{cases}
\exp\big(\langle \log L(j,k)\rangle\big), & q=0,\\
\langle L(j,k)^q\rangle, & q\neq 0,
\end{cases}
\]
where $\langle\cdot\rangle$ denotes averaging over spatial index $k$ at scale $j$.

Following the Chhabra--Jensen method, a scale-dependent probability measure $p_k(q,j)=L(j,k)^q / \sum_k L(j,k)^q$ was defined, from which the singularity exponent $h(q)$ and the multifractal spectrum $D(q)$ were estimated via linear regression across scales.
These quantities characterized the local scaling behavior and the geometry of the multifractal spectrum, respectively.

From the resulting multifractal scaling curves, a 16-dimensional multifractal descriptor was constructed for each channel, including:
(i) linear and quadratic coefficients obtained by fitting the scaling function with first- and second-order polynomials,
(ii) the dynamic range and selected reference values of the scaling function at specific moment orders, and
(iii) a set of singularity exponent samples extracted around the peak of the multifractal spectrum. These descriptors jointly capture global scaling trends and local spectrum geometry.

The resulting per-channel multifractal feature matrix was
\[
\mathbf{F}_{\mathrm{mf}}\in\mathbb{R}^{62\times16}.
\]



\subsection{The Dual-Branch Transformer Backbone (MAET)}

The channel-wise EEG features were concatenated and flattened into a 558-dimensional vector, while the global descriptors were aggregated into a 25-dimensional trial-level vector. These two feature vectors were taken as two input modalities of the proposed model.

We adopted a dual-branch transformer backbone (MAET) \cite{seedvii}. However, unlike~\cite{seedvii}, where MAET was employed to fuse EEG and eye-tracking features, our framework jointly processes \emph{local} and \emph{global} EEG representations (Fig.~\ref{fig:maet}). Specifically, the local EEG feature vector $\mathbf{eeg} \in \mathbb{R}^{558}$ and the global auxiliary feature vector $\mathbf{aux} \in \mathbb{R}^{25}$ were first projected into a shared embedding space via modality-specific multi-view embedding layers:
\[
    \mathbf{Z}_{\text{eeg}} = f_{\text{mv}}^{\text{eeg}}(\mathbf{eeg}), \qquad
    \mathbf{Z}_{\text{aux}} = f_{\text{mv}}^{\text{aux}}(\mathbf{aux}),
\]
where $f_{\text{mv}}^{(\cdot)}$ denotes a linear projection with learned gating and normalization.

The resulting embeddings were concatenated into a unified token sequence and processed by $L$ stacked transformer blocks. Each block consisted of multi-head self-attention followed by a feed-forward network with residual connections and layer normalization:
\[
    \mathbf{H}^{(l+1)} =
    \mathrm{LN}\big(
        \mathbf{H}^{(l)} +
        \mathrm{MHA}(\mathbf{H}^{(l)})
    \big).
\]

Through self-attention, information from the local and global features were adaptively integrated at the representation level, enabling the model to capture complementary local dynamics and global characteristics. The final hidden representation corresponding to the classification token was then fed into a single emotion classification head:
\[
    \hat{y} = \mathrm{softmax}(W_o \mathbf{h}_{\text{cls}} + b),
\]
where $\mathbf{h}_{\text{cls}}$ denotes the fused representation produced by the transformer backbone.

To improve cross-subject generalization, domain-adversarial training was employed during optimization. A gradient reversal layer (GRL) was applied to the shared representation to encourage subject-invariant feature learning, leading to the following training objective:
\[
    \mathcal{L} =
    \mathcal{L}_{\text{cls}} +
    \lambda \mathcal{L}_{\text{domain}}.
\]
Such representation-level regularization strategies were closely related to recent advances in learning subject-invariant EEG representations through contrastive learning and domain generalization techniques~\cite{ganin2016dann,shen2023contrastive,apicella2024generalization}.

\section{EXPERIMENTS AND RESULTS}

This section presents the quantitative evaluation of the proposed local--global connectivity fusion framework on the SEED-VII dataset under a leave-one-subject-out (LOSO) protocol. For emotion recognition performance, we report classification accuracy and macro-F1 score for the 7-class setting. In addition, confusion matrix analyses are provided to examine class-wise performance. Finally, ablation studies are presented to assess the contribution of each component of the proposed framework. The code has been released at \url{https://github.com/Danielz-z/LGF-EEG-Emotion}.

\subsection{Emotion Recognition Performance}

For each LOSO fold, one subject was held out as the test set while the remaining subjects form the training and validation sets.
Subject-wise normalization was computed on the training subjects and applied to both training and test samples. To mitigate distribution shifts, a fallback normalization statistic was computed as the average of all training-subject statistics. Final performance was reported as the mean across all LOSO folds.

Table~\ref{tab:main_7class} summarizes the cross-subject performance of the proposed method in comparison with representative baseline using only DE features. The proposed method achieved the highest performance, reaching approximately 40\% accuracy under the LOSO evaluation protocol. This result shows the effectiveness of jointly leveraging channel-wise EEG representations and trial-level global descriptors for subject-independent emotion recognition.

\begin{table}[t]
\centering
\caption{Performance of 7-class emotion recognition on the SEED-VII dataset under a LOSO protocol. For MAET variants, each fold was repeated with $R{=}3$ random seeds and averaged prior to computing the final mean across folds.}
\label{tab:main_7class}
\begin{tabular}{lcc}
\toprule
\textbf{Method} & \textbf{Acc. (\%)} & \textbf{Macro-F1 (\%)} \\
\midrule
MAET (EEG310, DE)  \cite{seedvii}              & 36.4 & 24.2 \\
\textbf{MAET (EEG558 + Aux25) (Ours)}    & \textbf{40.1} & \textbf{38.7} \\
\bottomrule
\end{tabular}
\end{table}

\begin{figure}[t]
    \centering
    \includegraphics[width=\columnwidth]{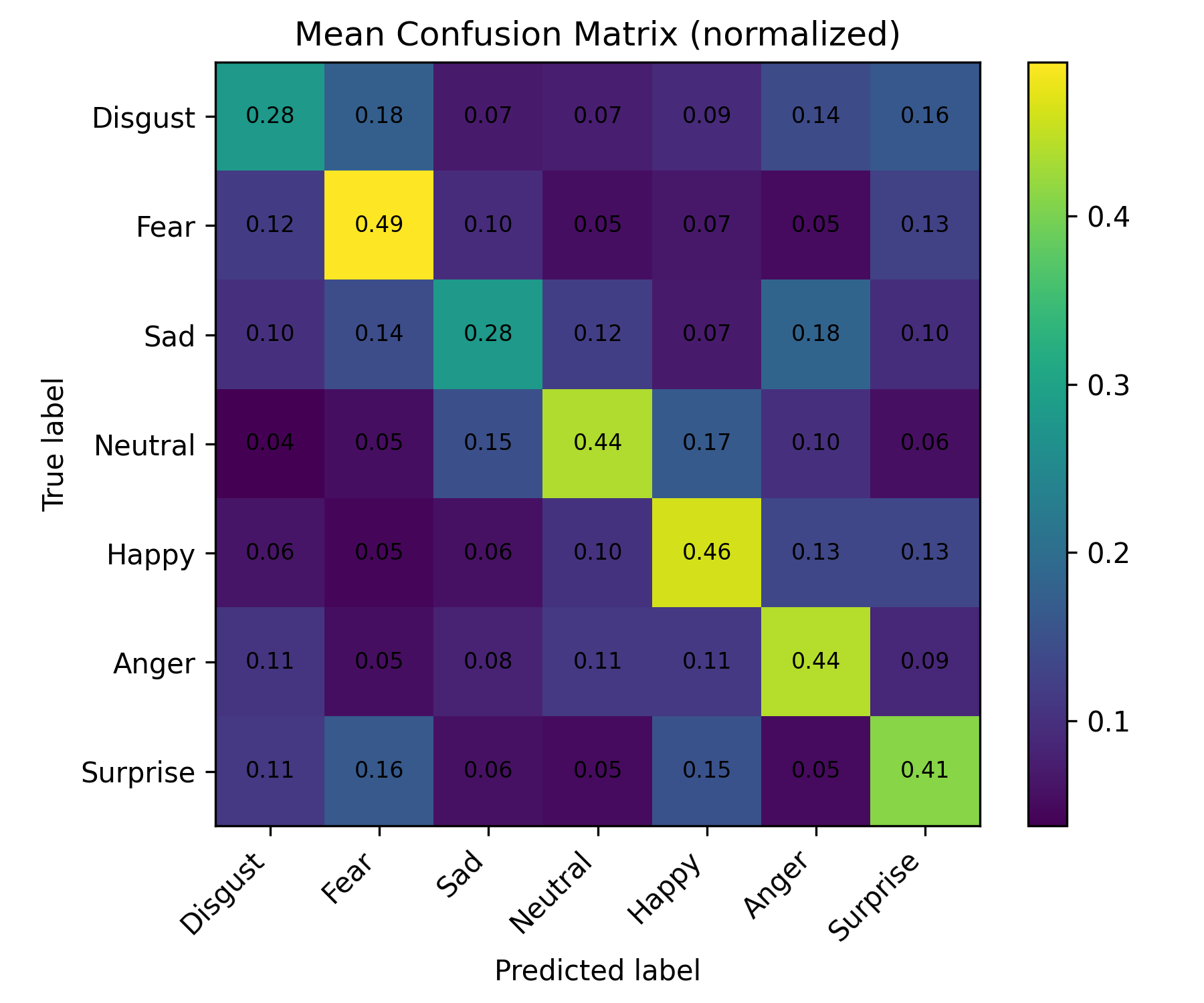}
    \caption{Mean normalized confusion matrix averaged over LOSO folds for 7-class emotion recognition.}
    \label{fig:confmat}
\end{figure}

Fig.~\ref{fig:confmat} shows the normalized confusion matrix averaged over LOSO folds. The proposed model achieves relatively higher recognition rates for high-arousal emotions such as fear and happiness, whereas emotions with lower intensity or overlapping neural patterns (e.g., disgust and sadness) remain more challenging. This observation is consistent with prior findings on SEED-family datasets~\cite{zheng2019tac,apicella2024generalization, niaki2025bipartite}.

\begin{figure}[t]
    \centering
    \includegraphics[width=\columnwidth]{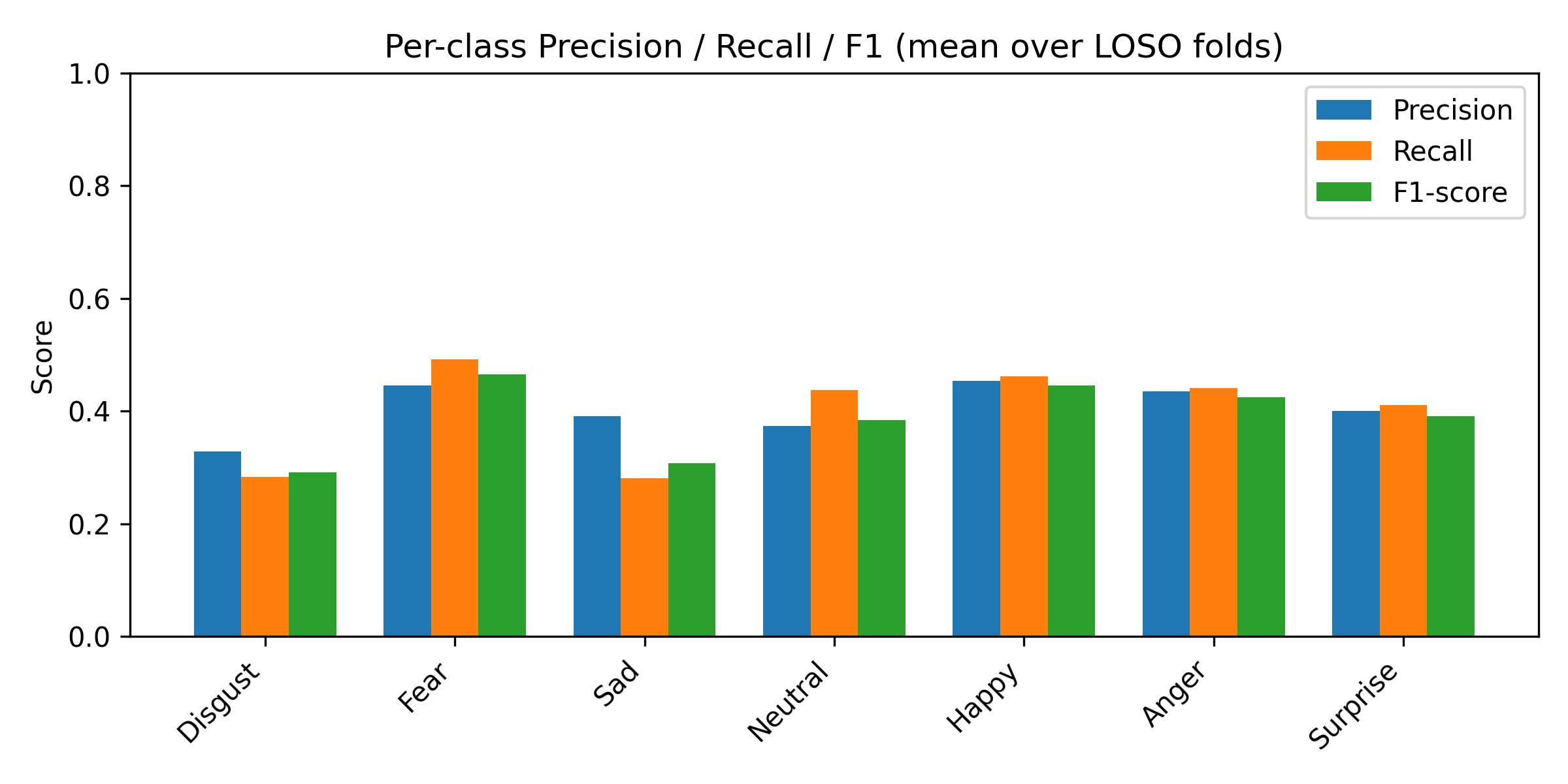}
    \caption{Per-class precision, recall, and F1-score (mean over LOSO folds).}
    \label{fig:perclass_prf}
\end{figure}

Fig.~\ref{fig:perclass_prf} presents the emotion recognition performance in terms of precision, recall, and F1-score. Notably, all emotions achieve performance levels higher than random chance ($1/7 = 14.3\%$).

\begin{figure}[t]
    \centering
    \includegraphics[width=\columnwidth]{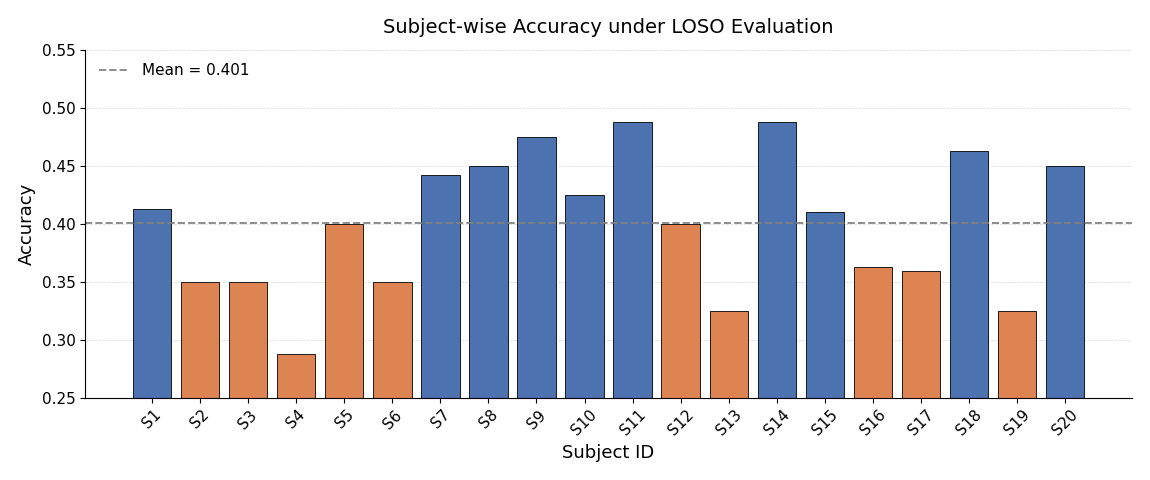}
    \caption{Subject-wise accuracy under LOSO evaluation. The dashed line denotes the mean accuracy across subjects.}
    \label{fig:subj_acc}
\end{figure}

Regarding subject-wise performance (Fig.~\ref{fig:subj_acc}), noticeable inter-subject variability remained. Nevertheless, the proposed model exhibited relatively stable performance across most subjects, suggesting that the integration of local and global features may contribute to improved robustness across participants.

\subsection{Ablation Studies}

To better understand the contribution of different feature components, we conducted a series of ablation experiments summarized in Table~\ref{tab:ablation_maet}. Using only the local EEG features (with a feature dimension of 558) resulted in a clear performance degradation compared with the full model, indicating that the inclusion of trial-level global features provided complementary information beyond local EEG representations.

We further evaluated the contribution of different global feature subsets. Among the individual global feature groups, incorporating only spectral features yielded the highest performance. This was followed by the multifractal descriptors, underscoring their effectiveness in capturing scale-dependent neural irregularities, which is consistent with prior findings on multifractal EEG analysis~\cite{eke2018fractal, wang2020mf}. In contrast, using only time-domain or fractal-domain global features resulted in more limited performance gains.

Finally, a simple early-fusion baseline based on direct feature concatenation performed worse than the proposed MAET framework, suggesting that representation-level integration via the transformer backbone was more effective than naive fusion strategies.

\begin{table}[t]
\centering
\caption{MAET ablation study under a LOSO protocol for 7-class emotion recognition.}
\label{tab:ablation_maet}
\setlength{\tabcolsep}{6pt}
\renewcommand{\arraystretch}{1.15}
\begin{tabular}{p{0.52\linewidth}cc}
\toprule
\textbf{Variant} & \textbf{Acc. (\%)} & \textbf{Macro-F1 (\%)} \\
\midrule
Local features only (558 dimension; no global features) 
    & 34.6 & 33.9 \\
Early fusion (only one branch concatenating all features: local and global, 558 + 25 = 583 features) 
    & 38.5 & 31.7 \\
Local features + global fractal features only
    & 37.2 & 36.0 \\
Local features + global time-domain features only features 
    & 36.8 & 35.9 \\
Local features +  global frequency-domain features only
    & 38.6 & 36.6 \\
\midrule
\textbf{Local features  + all global features} 
    & \textbf{40.1} & \textbf{38.7} \\
\bottomrule
\end{tabular}
\end{table}

\subsection{Ablation on Feature Groups Using Classical Classifiers}
Table~\ref{tab:baselines_full} reports the performance of classical machine learning baselines, including support vector machines (SVM), logistic regression (LogReg), multilayer perceptrons (MLP), and random forests (RF). Consistent with the MAET results, combining channel-wise local EEG features with global descriptors improved performance across all classifiers, confirming the complementary nature of the two feature groups.




\begin{table}[t]
\centering
\caption{Performance of classical machine learning baselines under a LOSO protocol for 7-class emotion recognition. Results are reported as mean $\pm$ standard deviation across folds.}
\label{tab:baselines_full}
\begin{tabular}{llcc}
\toprule
\textbf{Model} & \textbf{Feature Set} & \textbf{Acc. (\%)} & \textbf{Macro-F1 (\%)} \\
\midrule
SVM     & Local feat. only      & 21.7$\pm$4.2 & 15.8$\pm$5.7 \\
SVM     & Global feat. only      & 16.8$\pm$2.9 & 10.4$\pm$4.4 \\
SVM     & Local+global feat.      & 24.3$\pm$6.0 & 18.1$\pm$6.8 \\
\midrule
LogReg  & Local feat. only      & 21.0$\pm$4.4 & 14.9$\pm$5.6 \\
LogReg  & Global feat. only       & 16.0$\pm$2.7 &  9.8$\pm$4.0 \\
LogReg  &  Local+global feat.     & 23.8$\pm$6.1 & 17.6$\pm$6.9 \\
\midrule
MLP     & Local feat. only       & 22.5$\pm$5.2 & 16.6$\pm$6.3 \\
MLP     & Global feat. only      & 17.9$\pm$3.1 & 11.3$\pm$4.8 \\
MLP     &  Local+global feat.        & \textbf{25.7$\pm$6.3} & \textbf{20.3$\pm$7.3} \\
\midrule
RF      & Local feat. only   & 20.4$\pm$4.7 & 14.0$\pm$5.9 \\
RF      & Global feat. only     & 16.5$\pm$2.8 & 10.1$\pm$4.2 \\
RF      & Local+global feat.       & 23.1$\pm$5.9 & 17.0$\pm$6.6 \\
\bottomrule
\end{tabular}
\end{table}

\section{DISCUSSION}

Our results show that jointly modeling channel-wise EEG features and trial-level global descriptors leads to more robust subject-independent emotion recognition than using either component alone. This finding highlights the complementary roles of local neural dynamics and global coordination patterns in reducing inter-subject variability.

Channel-wise EEG representations derived from DE and graph-based features capture fine-grained electrode-specific activity relevant to emotion recognition. However, such local features are sensitive to inter-subject physiological differences and recording variability, limiting their generalization performance in cross-subject settings, consistent with prior studies~\cite{zheng2019tac,apicella2024generalization}.

To mitigate this limitation, the proposed framework incorporated trial-level global features that aggregate time-domain, spectral, and multifractal characteristics across channels. This aggregation reduces sensitivity to electrode-specific noise and session-related variability, yielding more stable representations that capture large-scale coordination patterns associated with affective processing~\cite{bullmore2009natrev,rubinov2010neuroimage}. Although insufficient on their own, these global features provide complementary contextual information when integrated with local EEG representations.

By fusing local and global features through a transformer-based backbone, the proposed framework enables adaptive interaction across feature scales and improves cross-subject generalization. These results suggest that effective affective state decoding benefits from jointly leveraging localized neural activity and subject-invariant coordination patterns.

\subsection{Subject Variability and Invariant Representation Learning}

Subject-independent emotion recognition remains challenging due to multiple sources of variability, including anatomical differences, electrode impedance, attention levels, emotional expressiveness, and session-related noise. In our experiments, subject-wise normalization combined with domain-adversarial training mitigates distribution shifts across participants, leading to more stable LOSO performance compared with classical baselines (see Table \ref{tab:baselines_full}).

We note, however, that a noticeable performance gap persists, indicating that learning fully subject-invariant representations from EEG signals remains an open problem. Future work may explore more advanced domain-invariant learning strategies or lightweight subject-adaptive calibration mechanisms to reduce subject dependence further.

\subsection{Contribution of Different Feature Types}

The ablation results highlight the importance of global features for subject-independent emotion recognition. Spectral and multifractal descriptors consistently improve performance, reflecting their ability to capture scale-dependent neural irregularities that are less sensitive to absolute signal amplitude. In contrast, time-domain and frequency-domain features provide more limited improvements when used in isolation.

\subsection{Limitations and Future Work}

Despite its effectiveness, the proposed framework exhibited some limitations. DE-based features compress spectral dynamics into predefined frequency bands, potentially obscuring finer-grained temporal patterns. The correlation-based connectivity matrix was estimated using a single temporal window and does not explicitly model time-varying functional connectivity, and the MAET backbone operates on flattened feature vectors without explicitly preserving spatial relationships among EEG channels.

Future work could address these limitations by incorporating graph-aware or spatiotemporal transformer architectures that operate directly on structured EEG representations. Extending the framework to jointly model discrete emotion categories and continuous affective dimensions, as well as integrating additional modalities such as eye tracking or facial expressions, may further improve robustness in real-world scenarios.

Overall, these findings underscore the importance of combining local neural dynamics with global trial-level characteristics to improve cross-subject generalization in EEG-based affective computing systems.

\section{CONCLUSION}
This work presented a local–global feature fusion framework for subject-independent EEG-based emotion recognition on the SEED-VII dataset. By jointly modeling channel-wise EEG features and trial-level global descriptors, the proposed approach captured complementary fine-grained neural dynamics and large-scale coordination patterns. The integration of a dual-branch transformer backbone with subject-wise normalization and domain-adversarial learning further enhanced cross-subject generalization. Experimental results under a LOSO protocol demonstrated that the proposed framework consistently outperformed local-only and classical baseline methods, achieving approximately 40\% accuracy in 7-class emotion recognition. These results underscore the importance of integrating local and global EEG representations for building robust and generalizable affective computing systems.



\bibliography{ref}
\bibliographystyle{IEEEtran}

\end{document}